\documentclass{article}

\usepackage{arxiv}

\usepackage[utf8]{inputenc} 
\usepackage[T1]{fontenc}    
\usepackage{authblk}
\usepackage{hyperref}       
\usepackage{url}            
\usepackage{booktabs}       
\usepackage{amsfonts}       
\usepackage{nicefrac}       
\usepackage{microtype}      
\usepackage{lipsum}
\usepackage{float}
\usepackage{amsmath}
\usepackage{fancyhdr}       
\usepackage{graphicx}       
\graphicspath{{media/}}     

\usepackage{multirow}
\usepackage[table,xcdraw]{xcolor}

\DeclareUnicodeCharacter{200B}{}

\title{Transformer based super-resolution downscaling for regional reanalysis: Full domain vs tiling approaches}

\author[1]{Antonio Pérez}
\author[1]{Mario Santa Cruz}
\author[1]{Daniel San Martín}
\author[2]{José Manuel Gutiérrez}

\affil[1]{Predictia Intelligent Data Solutions S.L. (Predictia), Santander, Spain}
\affil[2]{Instituto de Física de Cantabria (IFCA), CSIC-Universidad de Cantabria, Santander, Spain}

\begin{document}
\maketitle

\begin{abstract}

Super-resolution (SR) is a promising cost-effective downscaling methodology for producing high-resolution climate information from coarser counterparts. A particular application is downscaling regional reanalysis outputs (predictand) from the driving global counterparts (predictor). This study conducts an intercomparison of various SR downscaling methods focusing on temperature and using the CERRA reanalysis (5.5 km resolution, produced with a regional atmospheric model driven by ERA5) as example. The method proposed in this work is the Swin transformer and two alternative methods are used as benchmark (fully convolutional U-Net and convolutional and dense DeepESD) as well as the simple bicubic interpolation. We compare two approaches, the standard one using the \textit{full domain} as input and a more scalable \textit{tiling} approach, dividing the full domain into tiles that are used as input. The methods are trained to downscale CERRA surface temperature, based on temperature information from the driving ERA5; in addition, the tiling approach includes static orographic information. We show that the tiling approach, which requires spatial transferability, comes at the cost of a lower performance (although it outperforms some full-domain benchmarks), but provides an efficient scalable solution that allows SR reduction on a pan-European scale and is valuable for real-time applications.

\end{abstract}


\section{Introduction}
\label{sec:introduction}

Reanalysis datasets constitute the main source of spatially homogeneous information for climate analysis since they provide long records (spanning several decades) of physically consistent hourly/daily gridded data for many variables produced globally with a particular atmospheric general circulation model (AGCM) assimilating the available observations (see \url{https://reanalyses.org} for an overview of the current reanalyses). Besides the historical records, in some cases reanalyses provide near real-time information that allows monitoring the state of the climate. For instance, ERA5 \cite{hersbach2020era5} is the latest ECMWF climate reanalysis, providing hourly data on many atmospheric and land-surface parameters at 0.25º resolution, from 1940 to near real-time. However, much of this data is generated at coarse spatial resolutions, typically on the order of tens of kilometres, hampering their application for local and regional climate analysis, including extreme weather events, which often occur on smaller spatial scales. Enhancing the spatial resolution of reanalyses datasets is therefore critical for improving its utility for local-scale climate analysis and decision-making.

A number of downscaling methods have been developed over the last decades for improving the spatial resolution of AGCM outputs based on two main approaches \cite{maraun2017}: dynamical  and statistical downscaling. Dynamical downscaling employs regional atmospheric models (Limited Area Models, LAMs) over limited areas of interest, driven at the boundaries by the AGCM outputs, to increase their coarse-resolution. This approach allows to solve regional/local processes and provides physically consistent results, but is limited by its high computational demands. It has been recently applied to generate regional reanalysis over continental-wide areas, such as the CERRA renalysis over Europe using the HARMONIE-ALADIN regional model (driven by ERA5) at a 5.5km resolution.

Statistical downscaling is a data driven approach where  statistical relationships between coarse atmospheric variables (large-scale predictors) and high-resolution local-scale variables of interest are learnt from data (model simulations and observations). This approach is not very demanding and the recent use of deep learning techniques \cite{GoodBengCour16}, with their capacity to automatically learn complex spatiotemporal relationships from data, has given a big boost to its operationalisation in different downscaling applications, including super-resolution (SR) downscaling \cite{vandal2017deepsd}. SR was originally developed in the field of computer vision and has been successfully adapted for downscale weather and climate data using different deep learning techniques, from simple convolutional models to complex generative models (such GANs and diffusion models) allowing to estimate uncertainty  \cite{rampal2024enhancing}. The vision transformer (ViTs) models used in this work provide a promising avenue for SR downscaling by capturing long-range dependencies in data through attention mechanisms \cite{conde2022swin2sr}. They can effectively process large-scale climate data and have proven effective in tasks where both local and global spatial relationships are critical. 

Some of these SR downscaling methods have been recently applied to downscale global reanalysis products to regional scale, including convolutional \cite{Reddy_Matear_Taylor_Thatcher_Grose_2023} and diffusion models \cite{merizzi2024wind} for downscaling precipitation and wind, respectively. These models use data from existing global and regional reanalysis products over limited periods of time. The downscaling function learnt from data can be used, for instance, to produce real-time information from the global reanalyses with limited resources. However, SR reanalysis downscaling may require substantial computational resources for the training phase limiting their applicability to wide areas (e.g. continental-wide domains).

Here we present the results of a new SR downscaling method based on vision transformers and compare the standard full domain strategy with a new tiling implementation, dividing the domain into equally-sized spatial tiles and learning the downscaling function in the tile domain. This new implementation is computationally efficient and allows for cost-effective continental-wide applications, but requires spatial generalisation and transferability; in this case, static orographic information is added to the input to inform on the particular tile conditions allowing generalisation. This strategy builds on recent studies \cite{prasad2024evaluating} from the vision field demonstrating that utilising spatial patches can maintain a model's performance when applied to unseen areas. 
We focus on temperature and use the global and regional reanalysis ERA5 and CERRA over an area covering the Iberian peninsula, spanning a wide range of regional climate conditions. We used as benchmark a number of SR convolutional deep learning methods, including fully convolutional (UNet) and convolutional and dense (DeepESD). 

\section{Data and methods}
\label{sec:data_and_methods}

\subsection{Region of study}
\label{subsec:region-of-study}

Super-resolution (SR) models are highly sensitive to the size of the training domain due to the exponential increase in data points with larger geographical areas. As the domain size expands, the number of grid points increases exponentially, and this directly impacts the number of parameters required for the model. To mitigate these limitations, we focused  on a specific domain covering the Iberian Peninsula (see Figure \ref{fig:01}, top). This region presents a wide range of regional climates, from mountain to arid climates, along with pronounced orographic gradients and sea/land contrast regions. The reduced domain size allows the model to manage the increased resolution without the need for an overwhelming number of parameters, making it feasible to conduct high-resolution downscaling within a reasonable timeframe.

\begin{figure}[H]
    \centering
    \includegraphics[width=0.75\linewidth]{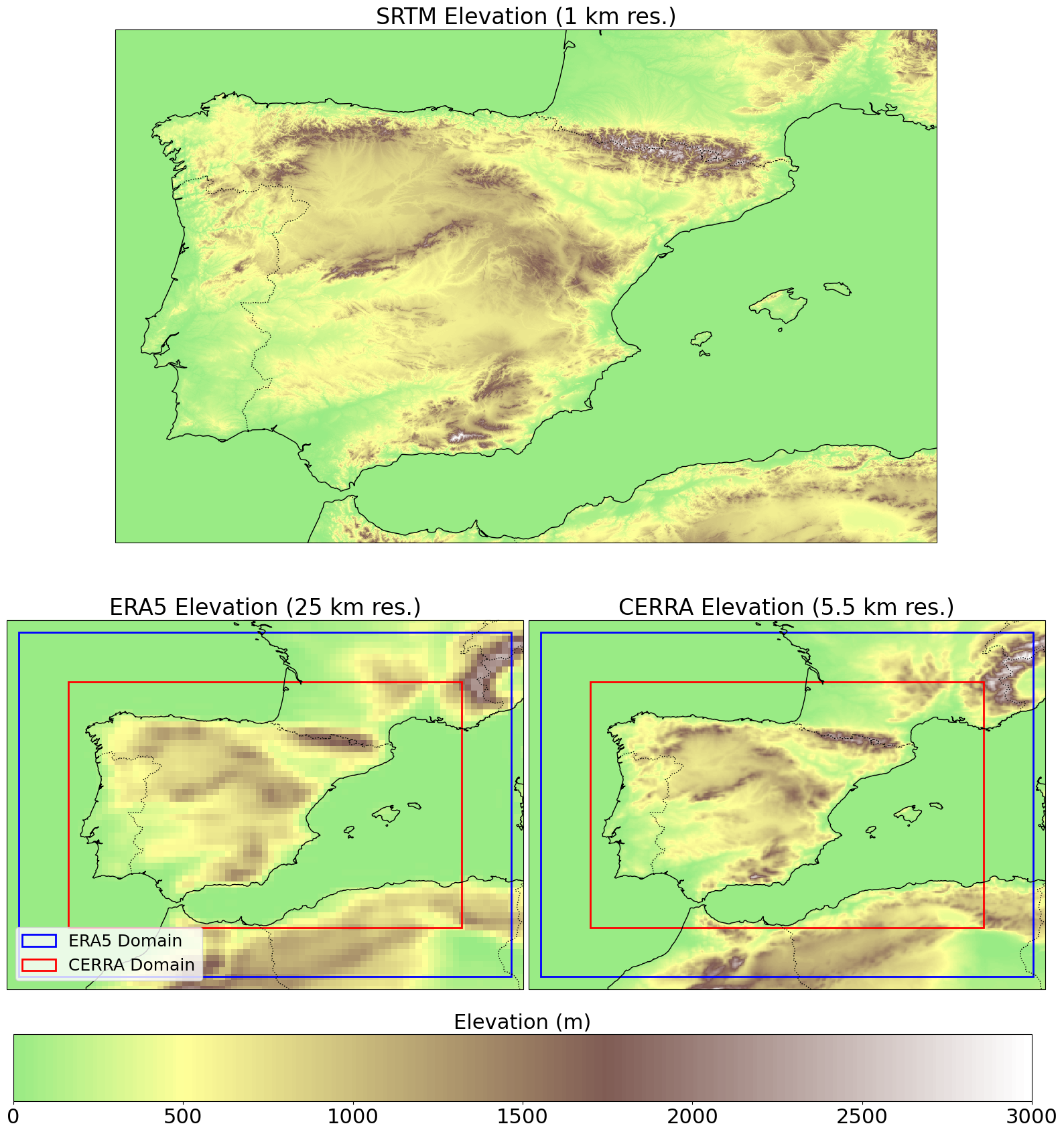}
    \caption{Spatial domain of the data sources used in this study. (top) High-resolution topography of the region of interest acquired from the NASA Shuttle Radar Topography Mission (SRTM). (bottom left) The ERA5 domain covers a range from 12°E to 8°W longitude and 33°N to 47°N latitude, with an elevation map at a spatial resolution of 0.25°. (bottom right) The CERRA domain spans from 10°E to 6°W longitude and 35°N to 45°N latitude, with an elevation map at a finer spatial resolution of 0.05°. The left panel shows the elevation from ERA5, while the right panel displays the higher-resolution elevation from CERRA.}
    \label{fig:01}
\end{figure}

\subsection{Reanalysis data: ERA5 and CERRA}
\label{subsec:era5-and-cerra}

ERA5 is the fifth generation ECMWF global reanalysis providing hourly gridded information from 1940 up to present, with a spatial resolution of about 25 kilometres (Figure \ref{fig:01}, bottom-left; \cite{hersbach2020era5}). ERA5 simulations are freely available from the Copernicus Climate Change Service (C3S) with a 5-day delay.

CERRA is a high-resolution, limited-area reanalysis for the European domain providing 3-hourly assimilated data from September 1984 to June 2021 at a horizontal resolution of 5.5 km (\ref{fig:01}, bottom-right). It is based on the data assimilation system of the HARMONIE-ALADIN regional model with lateral boundary conditions from ERA5 and uses large-scale constraints from the global reanalysis, ensuring consistency between global and regional atmospheric processes \cite{ridal2024}. CERRA simulations are freely available from the Copernicus Climate Change Service (C3S).  

In this work we consider 3-hourly gridded near surface temperature data from both reanalyses for the common period 1985-2020. We also use the static gridded elevation fields and the land/sea fraction from both reanalyses (see Figure \ref{fig:01}). 

\subsection{Swin vision transformer}
\label{subsec:swin-transformer}

The proposed model is based on the Swin v2 Transformer architecture \cite{conde2022swin2sr} \cite{liu2022swin}, specifically tailored for Super Resolution (SR) tasks due to its efficiency for handling high-resolution inputs and its ability to effectively model long-range dependencies within the data. The Swin v2 Transformer, an advanced version of the original Swin Transformer, operates by partitioning the input image into non-overlapping local windows, which are then processed using self-attention mechanisms. The windows are shifted between consecutive layers, allowing for cross-window connections that enhance the model's ability to capture global context. This design choice significantly reduces computational complexity while maintaining high performance, making it particularly well-suited for super-resolution tasks.

In the implementation presented in this paper (which we refer to as Swin2SR) the objective is to learn the residuals stemming from bicubic interpolation over ERA5. To this aim, our model incorporates upscaling and denoising blocks. The purpose of the upscaling block is to increase the spatial resolution of the intermediate feature maps to the desired high-resolution output. This block utilises techniques such as PixelShuffle to rearrange the low-resolution feature map into a higher-resolution space. After upscaling, the high-resolution images often contain noise and artefacts introduced during the upscaling process. The denoising block addresses this issue by applying sophisticated filtering techniques to remove noise and enhance the clarity of the super-resolved output. This block includes normalisation layers and additional Swin2SR processing stages to refine the high-resolution image.

\subsubsection{Standard approach: Full domain}
\label{subsec:standard_approach}

In the standard approach, the entire domain was used as the input and output for the super-resolution model (see Figure \ref{fig:01} for an schematic illustration). The input grids were structured as 4D tensors with ERA5 temperature values, with dimension [batch size, 1, 57, 81], where 57 and 81 represent the number of grid points along the latitude and longitude axes in ERA5, respectively. The outputs were represented as 4D tensors with dimension [batch size, 1, 200, 320], indicating the finer grid resolution compared to the input grids. However, this approach has a significant limitation: the model does not scale well with increased spatial coverage due to the substantially larger number of pixels in the image, making the training process computationally intensive and less efficient. 

\begin{figure}[H]
    \centering
    \includegraphics[width=0.75\linewidth]{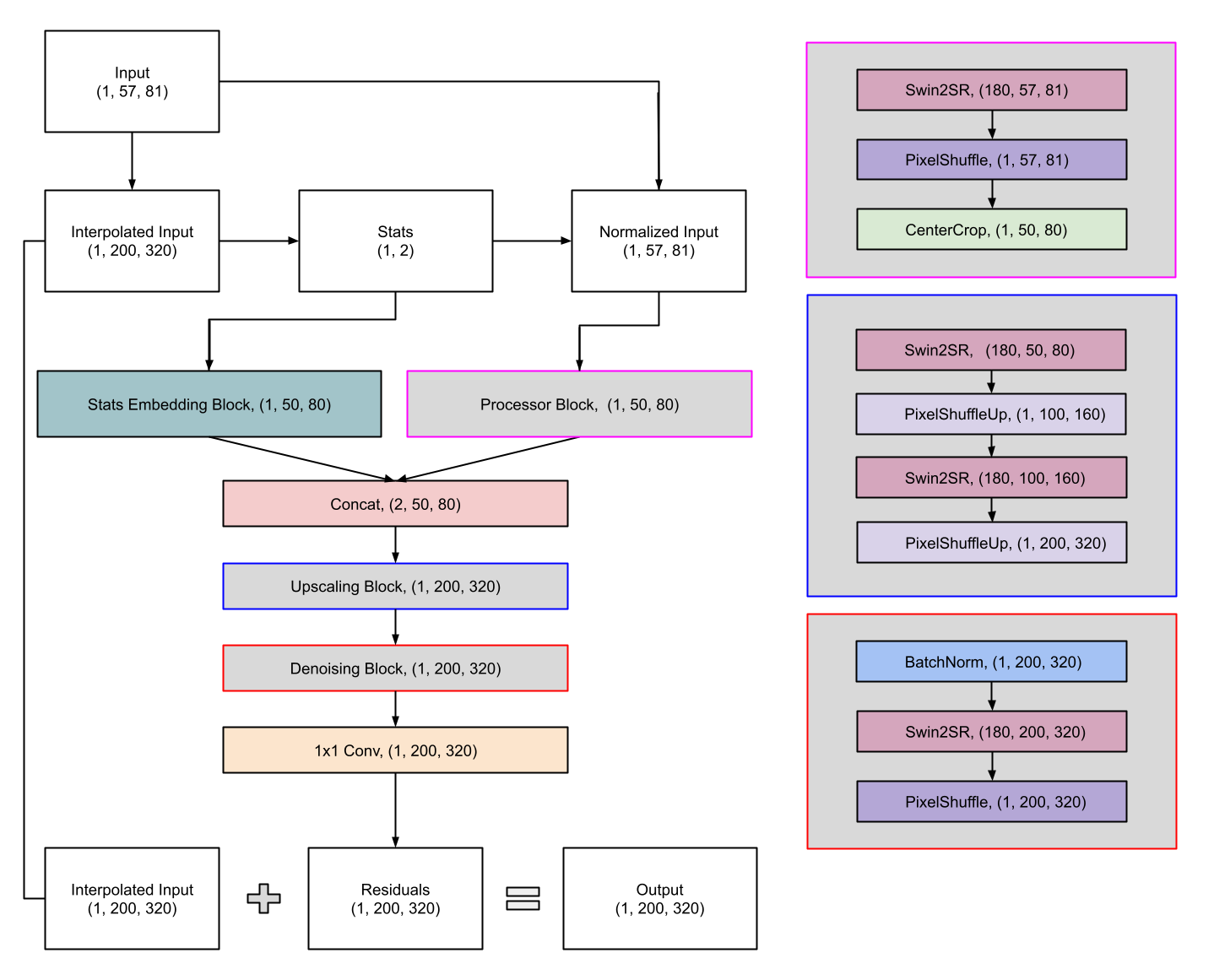}
    \caption{Diagram of the Swin2SR model for the full-domain approach illustrating the integration of the Swin v2 transformer model with a preprocessing module for upscaling spatial resolution. The full model upscales the processed input by a factor of 4, ensuring the output shape matches the region size of (200, 320).}
    \label{fig:01}
\end{figure}

\subsubsection{Proposed methodology: Tiling implementation}
\label{subsec:tiling_approach}

\begin{figure}[H]
    \centering
    \includegraphics[width=0.95\linewidth]{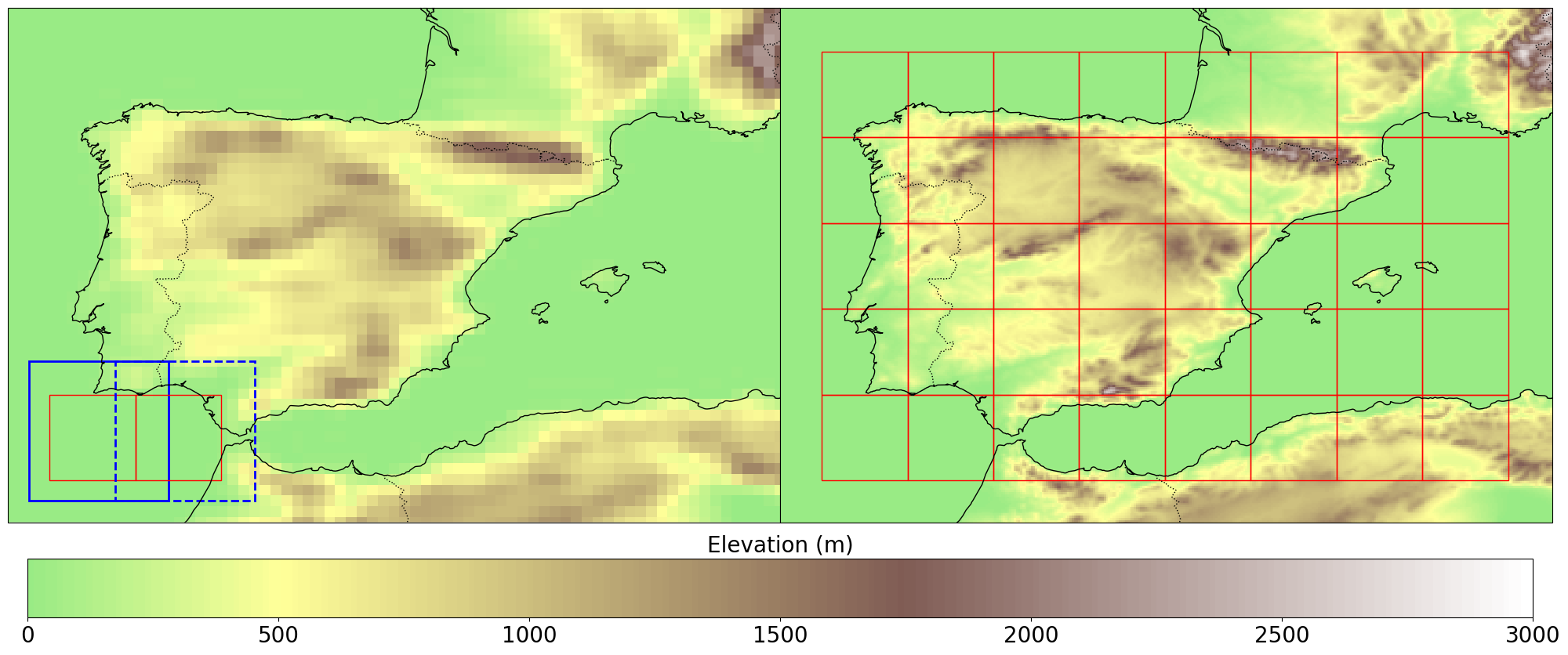}
    \caption{Visual representation of the data division into patches in the tiling implementation. On the left side, ERA5 orography data with an example of two different CERRA patches (red) surrounded by their corresponding ERA5 patch (blue) of size (13, 13). On the right side, CERRA orography with the full domain divided into 40 equal-size tiles.}
    \label{fig:03}
\end{figure}

\begin{figure}[H]
    \centering
    \includegraphics[width=1\linewidth]{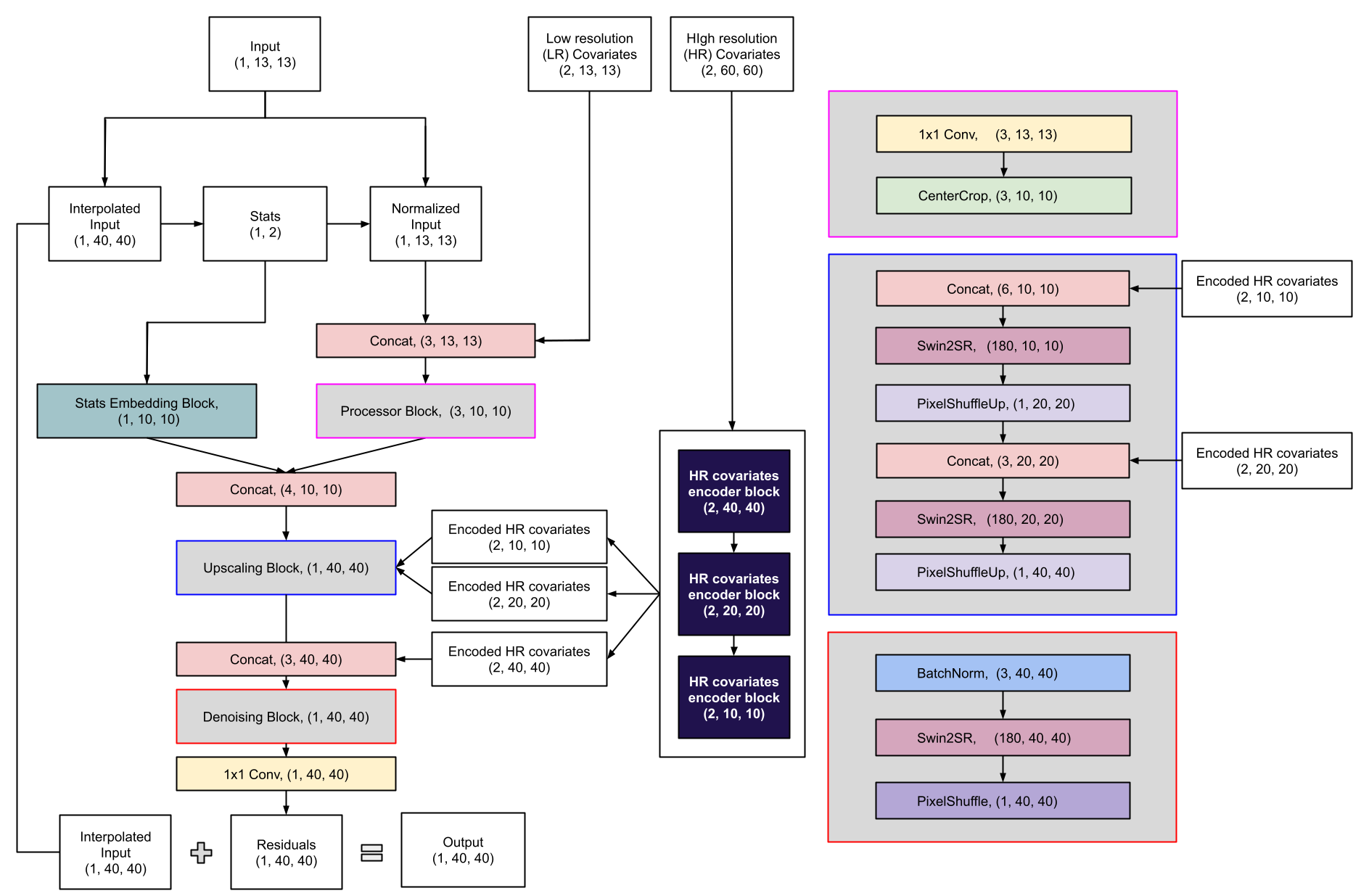}
    \caption{(Left) Schematic representation of the Swin2SR model for the patches approach, depicting the segmentation of the input data into smaller patches. The model upscales the processed patch by a factor of 4, ensuring an output shape of size (40, 40). (Right) Detailed view of the Swin2SR model blocks. The Processor Block (highlighted in pink) initiates the process with a 1x1 convolution followed by a centre crop, reducing the dimensions of the input data to a more manageable size for subsequent operations. The Upscaling Block (highlighted in blue) handles the enhancement of resolution, where concatenation with encoded high-resolution covariates is followed by a sequence of Swin2SR operations and pixel shuffling, progressively increasing the resolution from smaller patches to a final output size of 40x40 pixels. Finally, the Denoising Block (highlighted in red) further processes the upscaled data to reduce noise, applying batch normalisation, another Swin2SR operation, and a final pixel shuffle to ensure the output is both high-resolution and clean.}
    \label{fig:04}
\end{figure}

To mitigate the computational limitations hampering spatial scalability, we propose an alternative implementation wherein the output domain is divided into smaller spatial tiles (40 tiles of 40x40 grid points were selected) represented as 4D tensors with dimension [batch size, 1, 40, 40], corresponding to the finer grid resolution of the CERRA field, as shown in Figure \ref{fig:03} (right). In this case the input was structured as 4D tensors with dimension [batch size, 1, 13, 13], representing ERA5 temperature values (see Figure \ref{fig:04} for a schematic representation). A version of the tiling approach using overlapping tiles (patches) was also used to avoid artefacts in the tile boundaries; in this case each patch is randomly selected over the domain, avoiding boundary effects. 

The input tiles had a larger spatial coverage than the output tiles to provide a wider spatial context during model training; note that this leads to a partial overlapping as shown in Figure \ref{fig:03} (left). Additionally, several static covariates were provided as inputs to the model: the high- and low-resolution orography and land-sea masks from ERA5 and CERRA. Notably, the high-resolution covariates associated with a specific patch have a wider spatial coverage than the patch itself to provide more spatial context.

The low-resolution covariates are included in the model by concatenating them with the normalised input tensor. These concatenated inputs are then processed by the Processor Block, resulting in a tensor with dimension [3, 10, 10]. Meanwhile, the high-resolution covariates are processed through an encoder block, which encodes these covariates in three stages, producing outputs with dimensions [2, 10, 10], [2, 20, 20], and [2, 40, 40]. These encoded outputs provide crucial high-resolution information for the subsequent steps of the model. The concatenated and encoded tensors are then combined at various stages to ensure that both the low and high-resolution information is effectively utilised during the upscaling process.

\subsection{Benchmark models}
\label{subsec:methods-benchmark}

\subsubsection*{Bicubic interpolation}

Bicubic Interpolation is a widely used traditional technique for image upscaling \cite{keys1981cubic}. It works by using the weighted average of pixels within a 4x4 neighbourhood around each pixel to estimate new pixel values. This method is relatively simple and computationally efficient, making it a popular choice for quick image enhancement. However, while it can produce smoother images than nearest-neighbour or bilinear interpolation, bicubic Interpolation often fails to preserve fine details and sharp edges, leading to blurred or overly smoothed results, especially when dealing with high magnification factors.

\subsubsection*{UNet}

UNet is a convolutional neural network architecture originally developed for biomedical image segmentation \cite{ronneberger2015u}, but it has proven highly effective in super-resolution tasks as well. The UNet architecture consists of an encoder-decoder structure with skip connections that allow detailed spatial information to flow directly from the downsampling layers to the upsampling layers. This design enables UNet to capture both global context and fine-grained details, making it particularly effective in generating high-quality, high-resolution images from low-resolution inputs. Its ability to preserve structural details while enhancing resolution has made UNet a popular choice in various image processing applications.

\begin{figure}[H]
    \centering
    \includegraphics[width=1.02\linewidth]{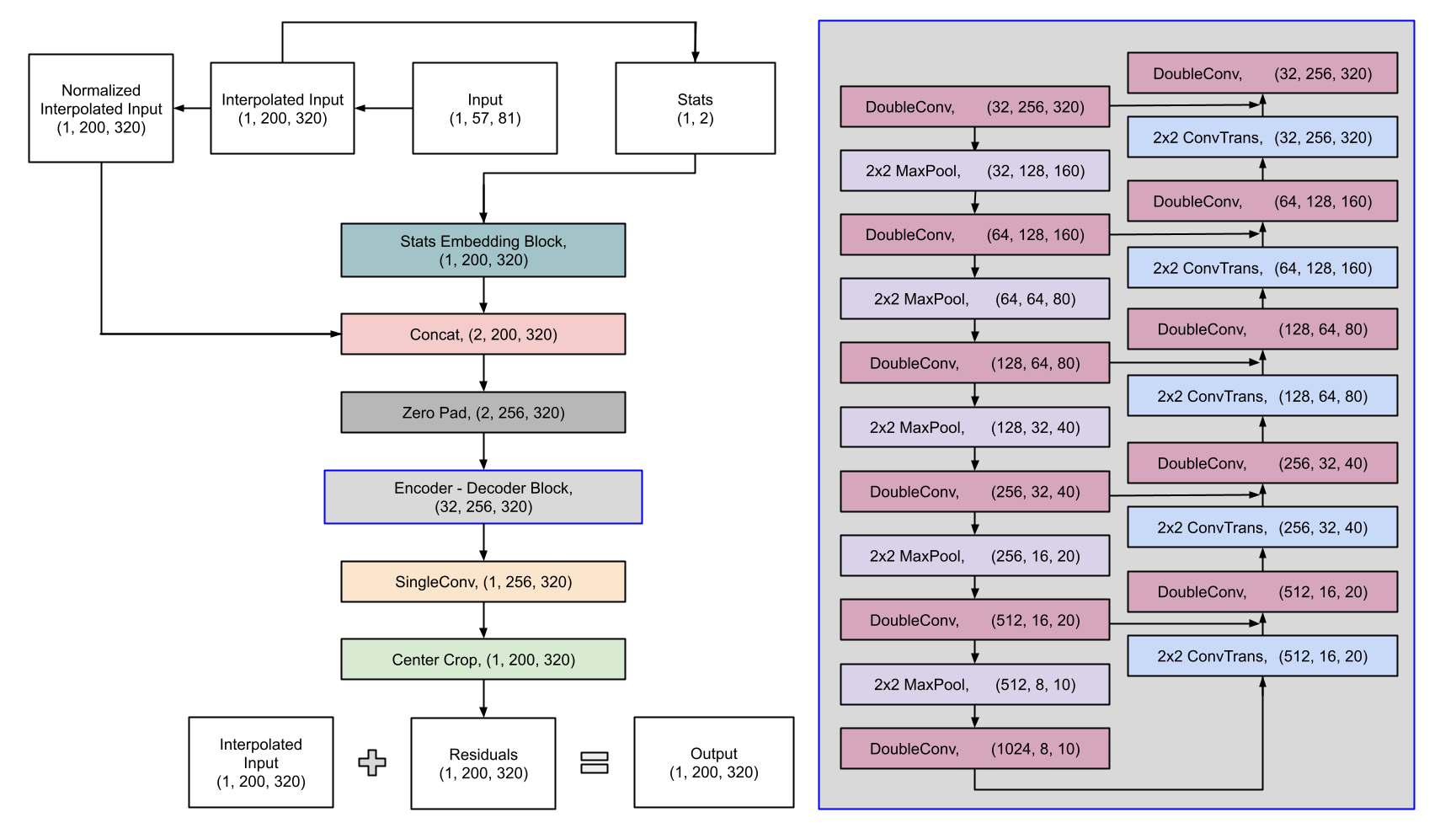}
    \caption{Architecture of the UNet model and detailed representation of its encoder-decoder block.}
    \label{fig:05}
\end{figure}

As shown in Figure \ref{fig:05}, the process begins with the normalised interpolated input and associated statistical data, which are combined in the Stats Embedding Block. The output from this block is concatenated with the original input and then zero-padded to ensure the dimensions are consistent for subsequent operations. The core of the model consists of an Encoder-Decoder Block, which captures both high-level and fine-grained features of the input data. 

The encoder module progressively reduces the spatial dimensions of the input data while increasing the depth, capturing essential features through a series of DoubleConv and MaxPooling layers. The decoder module then reconstructs the data back to the original dimensions using transposed convolutional layers (ConvTrans) that expand the spatial dimensions while reducing the depth. Skip connections between corresponding layers in the encoder and decoder help retain spatial information and improve the reconstruction quality. The final output of this block is a feature map that retains the high-resolution details necessary for accurate super-resolution.

After passing through this block, the data undergoes a single convolution operation and is then centre-cropped to match the desired output dimensions. The final high-resolution output is generated by adding the residuals obtained from the processed data to the interpolated input.

\subsubsection*{DeepESD}

DeepESD (Deep Empirical Statistical Downscaling) is an advanced deep learning model specifically designed for super-resolution tasks, particularly in the context of climate and environmental data \cite{banomedina2020}. DeepESD leverages a large-scale, high-capacity neural network to learn complex mappings from low-resolution to high-resolution data. With its deep architecture and large number of parameters, DeepESD excels in preserving fine details and reducing errors, outperforming traditional methods like Bicubic Interpolation. The model's strength lies in its ability to handle large datasets and deliver precise super-resolution outputs, making it a robust choice for scientific and environmental applications.

\begin{figure}[H]
    \centering
    \includegraphics[width=0.6\linewidth]{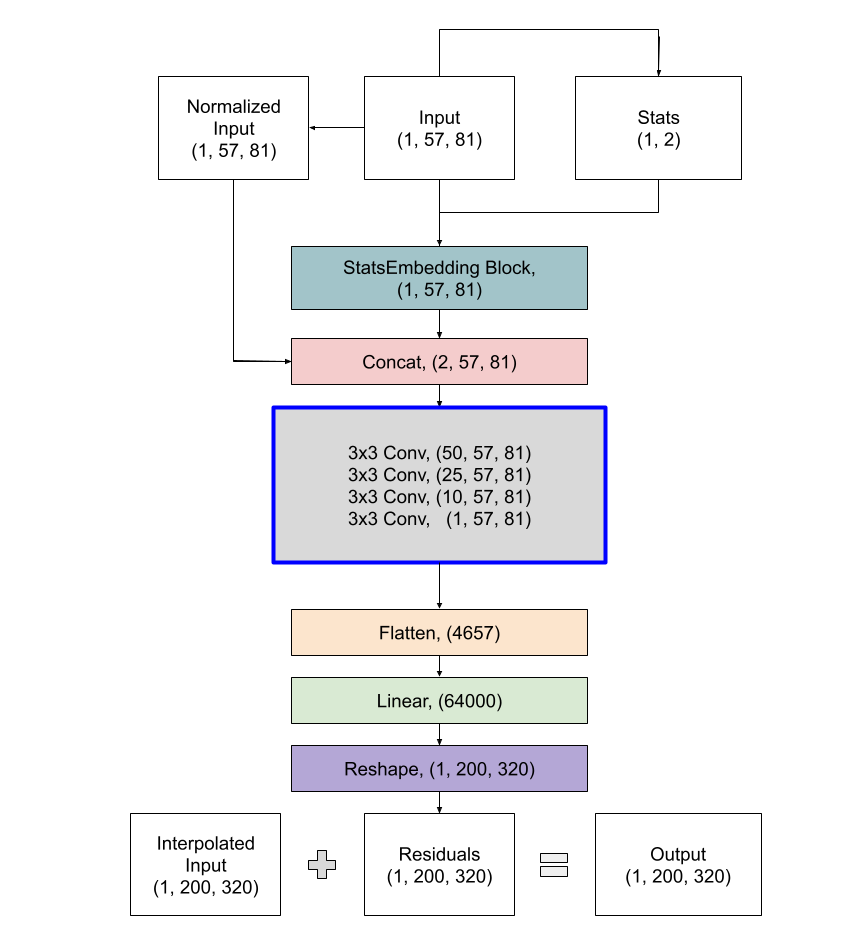}
    \caption{Architecture of the DeepESD Model.}
    \label{fig:06}
\end{figure}

As illustrated in Figure \ref{fig:06}, the process begins with the normalised input and statistical data, which are combined in the Stats Embedding Block. The concatenated output is then processed through a series of 3x3 convolutional layers with varying filter sizes, which gradually reduce the depth of the feature maps. After the convolutional operations, the data is flattened and passed through a linear layer, before being reshaped to match the dimensions of the final output. The final output is generated by adding the residuals from the processed data to the interpolated input, producing a high-resolution output.

\subsection{Training configuration}
\label{subsec:methods-train-config}

In this section, we outline several of the configuration options followed during training, including sampling strategies, normalisation, and the loss function used to optimise model performance.

The temporal division of the dataset is structured to optimise model training and subsequent epoch-based tests. The training period spans 29 years, from 1985 to 2013 and the internal test covers from 2014 to 2018; this five-year interval is dedicated to evaluating the model's performance on validation data to define early stopping during training (when the loss in the validation set is not improved during ten epochs by at least a 1$\%$). The independent period including 2019 and 2020 is used to evaluate the resulting models in Section 4.

For the standard approach, uniform sampling is performed, where all data samples have the same weight. In contrast, in the tiling approach, a specialised sampler was employed for the data loader to enhance the representativeness of samples during model training \cite{harris2022generative}. This sampler assigns greater weight to samples corresponding to patches with higher orographic variability and land-sea variability. The weight for each spatial patch​ is calculated as:

\begin{equation}
w_i = \left( \sum_{j=1}^{n} \sigma_{kj} \right) + \left( 1 - \frac{1}{N} \sum_{k=1}^{N} \sum_{j=1}^{n} \sigma_{kj} \right)
\end{equation}

where ${\sigma}_{kj}$ is the standard deviation of the j-th covariable for patch ${p}_{k}$, $n$ is the number of covariables, and $N$ is the total number of patches. The first term represents the variability of the current patch, while the second term serves to normalise and adjust the weight based on the average variability across all patches. This strategy ensures that the model receives an adequate number of examples from regions with complex and varied characteristics, which is crucial for improving the model's ability to generalise and make accurate predictions in these areas. By prioritising these more challenging patches, the model can better learn the subtle differences and specific patterns present in high-variability zones, resulting in a significant improvement in the resolution and accuracy of the generated outputs.

In both approaches, instance normalisation was performed. This is an alternative standardisation method that does not require historical data. This process involves normalising each sample independently by the mean and standard deviation of the input field, since the outputs are not available during inference. Two variations of this method were considered:

\begin{itemize}
    \item Using ERA5 statistics, which cover a wider area than CERRA.
    \item Using bicubic downscaled ERA5 statistics, which represent the same area as CERRA.
\end{itemize}

The difference between these two approaches lies in the fact that the input patch in the first method encompasses a larger area. Consequently, the second approach is deemed more appropriate, as the downscaled area distribution would be more similar to the output distribution.
After calculating the statistics (mean and standard deviation) for the instance normalisation approach, the data is standardised, and these statistics are embedded as a new channel. The two statistics are transformed through a linear layer to get a tensor with size equal to the multiplication of the elements of the input shape tensor. Afterwards, a reshaping layer is used to put it in a shape that permits the concatenation as the next step. This embedding process allows the model to effectively utilise the statistical information during training.

Finally, the SR model optimises its parameters using the Charbonnier loss function \cite{charbonnier1994loss}. The use of this function to enhance predictive accuracy and robustness comes from recent studies in the field of climate downscaling \cite{zhong2023investigating}. The Charbonnier loss, a differentiable variant of the L1 loss, helps improve the model's performance by providing a smoother approximation of the absolute error. This loss function is particularly effective in handling outliers and reducing the impact of noise, leading to more stable predictions. 

The Charbonnier loss function is defined as:

\begin{equation}
L_{\text{Charbonnier}}(x, y) = \sqrt{(x - y)^2 + \epsilon^2} 
\end{equation}

where $x$ represents the model predictions, $y$ represents the reference values or ground truth, and $\epsilon$ is a small constant added for numerical stability.

All SR approaches described in this paper were trained over the area of study on a V100 GPU with 16GB of memory, following the same training configuration above described with training times ranging from 10 hours (UNet and DeepESD) to 24 hours (SwinSR tiling) and to 72 hours (Swin2SR full-domain). 

\subsection{Evaluation indices}
\label{subsec:methods-evaluation}

In this study, we use several evaluation indices to assess the performance and accuracy of our model, focusing also on spatial representativity. These indices include Root Mean Squared Error (RMSE), Mean Absolute Error (MAE), Bias, Structural Similarity Index Measure (SSIM), and Peak Signal-to-Noise Ratio (PSNR). Each of these metrics provides a different perspective on the model's performance.

\subsubsection*{Root Mean Squared Error (RMSE)}
RMSE is calculated as the square root of the average of squared differences between prediction and actual observation. RMSE is defined as:

\begin{equation}
    \text{RMSE} = \sqrt{\frac{1}{N} \sum_{i=1}^{N} (y_i - \hat{y}_i)^2}
\end{equation}

where \( y_i \) is the actual value, \( \hat{y}_i \) is the predicted value, and \( N \) is the number of observations.

\subsubsection*{Mean Absolute Error (MAE)}
The MAE measures the average magnitude of absolute errors between predicted and observed values. It provides a straightforward indication of the prediction accuracy by taking the absolute differences. The formula for MAE is:

\begin{equation}
    \text{MAE} = \frac{1}{N} \sum_{i=1}^{N} \left| y_i - \hat{y}_i \right|
\end{equation}

where \( y_i \) is the actual value, \( \hat{y}_i \) is the predicted value, and \( N \) is the number of observations. Unlike RMSE, MAE gives equal weight to all errors, making it less sensitive to outliers.

\subsubsection*{Bias}
Bias is a metric used to quantify the systematic error in the predictions. It refers to the average difference between predicted and actual values, indicating whether predictions tend to overestimate or underestimate the true values. The formula for bias is:

\begin{equation}
    \text{Bias} = \frac{1}{N} \sum_{i=1}^{N} (\hat{y}_i - y_i)
\end{equation}

where \( y_i \) is the actual value, \( \hat{y}_i \) is the predicted value, and \( N \) is the number of observations. A positive bias indicates overestimation, while a negative bias indicates underestimation. Bias helps to assess whether the model systematically deviates from the true values.

\subsubsection*{Structural Similarity Index Measure (SSIM)}
SSIM is used to measure the similarity between two images. It considers changes in structural information, luminance, and contrast. The SSIM index is defined as:

\begin{equation}
    \text{SSIM} = \frac{(2\mu_y \mu_{\hat{y}} + C_1)(2\sigma_{y\hat{y}} + C_2)}{(\mu_y^2 + \mu_{\hat{y}}^2 + C_1)(\sigma_y^2 + \sigma_{\hat{y}}^2 + C_2)}
\end{equation}

where \( \mu_y \) and \( \mu_{\hat{y}} \) are the average of \( y \) and \( \hat{y} \), \( \sigma_y^2 \) and \( \sigma_{\hat{y}}^2 \) are the variances of \( y \) and \( \hat{y} \), \( \sigma_{y\hat{y}} \) is the covariance of \( y \) and \( \hat{y} \), and \( C_1 \) and \( C_2 \) are constants to stabilise the division.

\subsubsection*{Peak Signal-to-Noise Ratio (PSNR)}
PSNR is used to measure the quality of a reconstructed signal compared to its original version. It is defined as:

\begin{equation}
    \text{PSNR} = 10 \cdot \log_{10}\left(\frac{\text{MAX}^2}{\text{MSE}}\right)
\end{equation}

where \(\text{MAX}\) is the maximum possible pixel value of the image. For an 8-bit image, \(\text{MAX}\) is 255. \(\text{MSE}\) is the Mean Squared Error between the original and reconstructed image.

\section{Results}
\label{sec:results}
The performance of the different SR methods trained to downscale temperature over the full domain is shown in Table \ref{tab:performance_metrics_full_domain}; the table shows the spatial mean values of gridbox results for the different evaluation metrics described in Sec. \ref{subsec:methods-evaluation}. As expected, the bicubic interpolation significantly underperforms compared to the different SR methods, with Swin2SR exhibiting the best results for all metrics and seasons. When evaluating the overall accuracy of the methods, Swin2SR consistently achieves the best results with an annual average RMSE of 0.92, a notable improvement over DeepESD (1.00) and UNet (0.96), and far superior to the simple bicubic interpolation (1.30); similar results hold for MAE, with a mean annual value of 0.68 for Swin2SR and 0.72 and 0.75 for UNet and DeepESD, respectively. UNet achieves the best result for the bias with quite small values for Swin2SR and DeepESD as well. In terms of structural similarity, which is crucial for preserving the spatial patterns of temperature, Swin2SR outperforms the other models across all seasons (note that higher values of these indices indicate better performance), although UNet achieves comparable results. Overall, the performance of DeepESD is far from the other two SR methods, which exhibit a similar overall performance.

\begin{table}[h!]
\centering
\begin{tabular}{|cr|ccccc|}
\hline
\rowcolor[HTML]{EFEFEF} 
\multicolumn{1}{|l}{\cellcolor[HTML]{EFEFEF}}                                 & \multicolumn{1}{l|}{\cellcolor[HTML]{EFEFEF}} & \textbf{DJF}   & \textbf{MAM}   & \textbf{JJA}   & \textbf{SON}   & \textbf{Annual} \\ \hline
\multicolumn{1}{|c|}{}                                                        & \textbf{Bicubic Interp.}                      & 1,33           & 1,26           & 1,31           & 1,26           & 1,30            \\
\multicolumn{1}{|c|}{}                                                        & \textbf{DeepESD}                              & 1,03           & 0,96           & 1,02           & 0,98           & 1,00            \\
\multicolumn{1}{|c|}{}                                                        & \textbf{Swin2SR}                              & \textbf{0,91}  & \textbf{0,89}  & \textbf{0,96}  & \textbf{0,89}  & \textbf{0,92}   \\
\multicolumn{1}{|c|}{\multirow{-4}{*}{\textbf{RMSE}}}                         & \textbf{UNet}                                 & 0,97           & 0,92           & 1,00           & 0,94           & 0,96            \\ \hline
\rowcolor[HTML]{EFEFEF} 
\multicolumn{1}{|c|}{\cellcolor[HTML]{EFEFEF}}                                & \textbf{Bicubic Interp.}                      & 1,04           & 0,99           & 1,03           & 0,99           & 1,01            \\
\rowcolor[HTML]{EFEFEF} 
\multicolumn{1}{|c|}{\cellcolor[HTML]{EFEFEF}}                                & \textbf{DeepESD}                              & 0,77           & 0,72           & 0,77           & 0,74           & 0,75            \\
\rowcolor[HTML]{EFEFEF} 
\multicolumn{1}{|c|}{\cellcolor[HTML]{EFEFEF}}                                & \textbf{Swin2SR}                              & \textbf{0,68}  & \textbf{0,66}  & \textbf{0,72}  & \textbf{0,67}  & \textbf{0,68}   \\
\rowcolor[HTML]{EFEFEF} 
\multicolumn{1}{|c|}{\multirow{-4}{*}{\cellcolor[HTML]{EFEFEF}\textbf{MAE}}}  & \textbf{UNet}                                 & 0,72           & 0,69           & 0,75           & 0,71           & 0,72            \\ \hline
\multicolumn{1}{|c|}{}                                                        & \textbf{Bicubic Interp.}                      & 0,24           & 0,28           & 0,26           & 0,28           & 0,26            \\
\multicolumn{1}{|c|}{}                                                        & \textbf{DeepESD}                              & -0,01          & -0,05          & -0,08          & \textbf{0,00}  & -0,04           \\
\multicolumn{1}{|c|}{}                                                        & \textbf{Swin2SR}                              & -0,02          & -0,02          & -0,10          & -0,01          & -0,04           \\
\multicolumn{1}{|c|}{\multirow{-4}{*}{\textbf{Bias}}}                         & \textbf{UNet}                                 & \textbf{0,00}  & \textbf{-0,01} & \textbf{-0,06} & \textbf{0,00}  & \textbf{-0,02}  \\ \cline{1-3}
\rowcolor[HTML]{EFEFEF} 
\multicolumn{1}{|c|}{\cellcolor[HTML]{EFEFEF}}                                & \textbf{Bicubic Interp.}                      & 0,68           & 0,70           & 0,70           & 0,71           & 0,70            \\
\rowcolor[HTML]{EFEFEF} 
\multicolumn{1}{|c|}{\cellcolor[HTML]{EFEFEF}}                                & \textbf{DeepESD}                              & 0,83           & 0,88           & 0,88           & 0,87           & 0,87            \\
\rowcolor[HTML]{EFEFEF} 
\multicolumn{1}{|c|}{\cellcolor[HTML]{EFEFEF}}                                & \textbf{Swin2SR}                              & \textbf{0,87}  & \textbf{0,91}  & \textbf{0,90}  & \textbf{0,90}  & \textbf{0,89}   \\
\rowcolor[HTML]{EFEFEF} 
\multicolumn{1}{|c|}{\multirow{-4}{*}{\cellcolor[HTML]{EFEFEF}\textbf{SSIM}}} & \textbf{UNet}                                 & 0,86           & 0,90           & \textbf{0,90}  & 0,89           & \textbf{0,89}   \\ \hline
\multicolumn{1}{|c|}{}                                                        & \textbf{Bicubic Interp.}                      & 23,47          & 23,47          & 23,80          & 23,70          & 23,61           \\
\multicolumn{1}{|c|}{}                                                        & \textbf{DeepESD}                              & 27,79          & 28,59          & 28,25          & 28,17          & 28,20           \\
\multicolumn{1}{|c|}{}                                                        & \textbf{Swin2SR}                              & \textbf{28,93} & \textbf{29,38} & \textbf{28,97} & \textbf{29,03} & \textbf{29,08}  \\
\multicolumn{1}{|c|}{\multirow{-4}{*}{\textbf{PSNR}}}                         & \textbf{UNet}                                 & 28,50          & 29,18          & 28,70          & 28,67          & 28,76           \\ \hline
\end{tabular}

\label{tab:performance_metrics_full_domain}

\caption{Performance comparison of super-resolution methods for the full-domain approach across seasonal and annual metrics. This table presents the RMSE, MAE, Bias, SSIM, and PSNR values for different super-resolution methods, including Bicubic Interpolation, DeepESD (~300M parameters), Swin2SR transformer (~30M parameters), and UNet (~30M parameters), evaluated across different seasons (DJF, JJA, MAM, SON) and annually.}
\end{table}

\begin{figure}[H]
    \centering
    \includegraphics[width=0.92\linewidth]{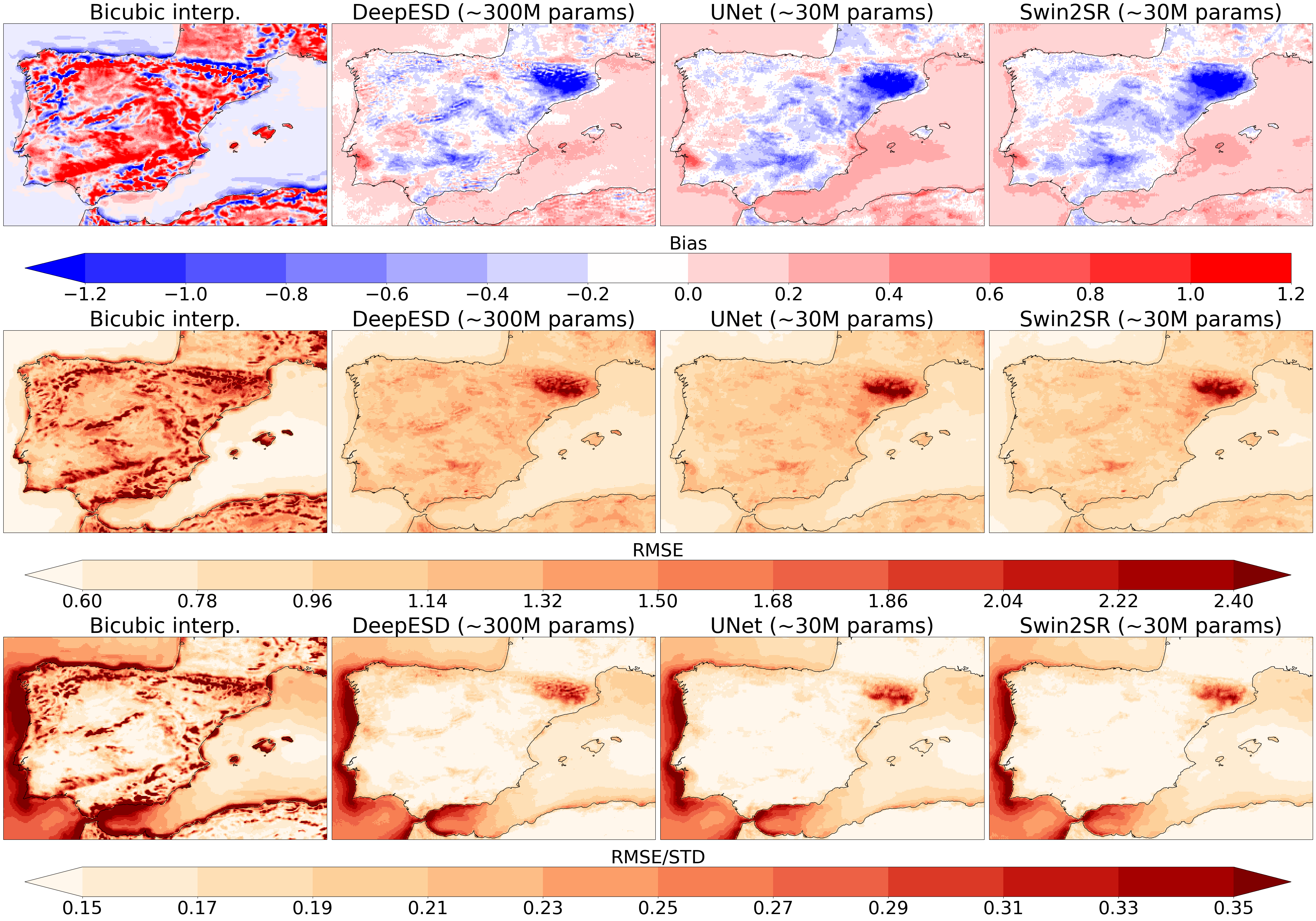}
    \caption{Spatial evaluation results for the different  SR downscaling methods for the full-domain approach, with the methods represented in columns, Bicubic interpolation (simple benchmark), Swin2SR transformer (~30M parameters), DeepESD (~300M parameters) and UNet (~30M parameters), from left to right, and evaluation metrics in rows (bias, RMSE, and RMSE/STD, from top to bottom). }
    \label{fig:07}
\end{figure}

Figure \ref{fig:07} shows the spatial details of the evaluation metrics. As expected, the errors for the bicubic interpolation are particularly large in regions of complex orography with steep orographic gradients and in coastal regions. In contrast, the three deep learning methods (UNet, DeepESD, and Swin2SR) exhibit very similar spatial patterns across the entire study domain. The largest biases and errors for the three SR methods are consistently observed in Southeastern Pyrenees, with similar and even larger errors locally than the bicubic interpolation benchmark. While this is a region of orographic contrast, with low and high elevations, it is not particularly different from other regions and we have no explanation for this behaviour caused by the systematic negative bias exhibited by the models, underestimating temperature values.

\begin{figure}[H]
    \centering
    \includegraphics[width=0.85\linewidth]{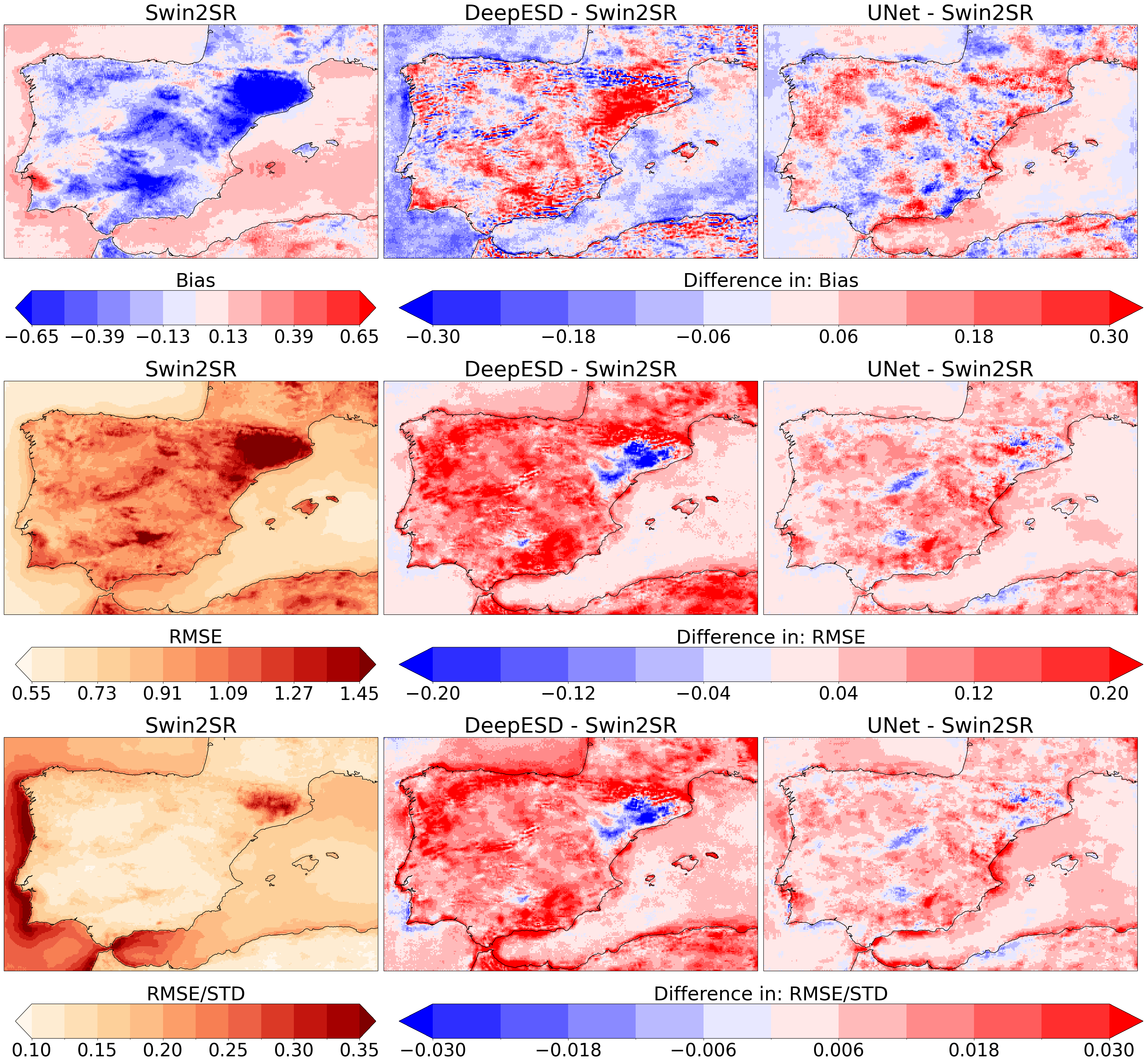}
    \caption{As Figure \ref{fig:07} but displaying the evaluation metrics for the best performing method (Swin2SR) and the differences with other super-resolution methods (in columns Swin2SR transformer, DeepESD and UNet, from left to right) for the full-domain approach. Different evaluation metrics are displayed in rows: bias, RMSE, and RMSE/STD, from top to bottom. The left column shows the actual values of the metrics for the Swin2SR method, whereas the second and third show differences of DeepESD and UNet relative to Swen2SR.}
    \label{fig:08}
\end{figure}

Figure \ref{fig:08} provides further insight in this problem, showing that the biases and errors are larger for UNet and Swin2SR methods, than for DeepESD. This figure compares the evaluation results of the UNet and DeepESD methods with those resulting from the best performing model, Swin2SR. The biases of the DeepESD method over the Southeastern Pyrenees region, although still large (see Figure \ref{fig:07}), are smaller than for the other methods, which could be an indication of a different relationship between ERA5 and CERRA over this area; note that the final dense layer of DeepESD provides individual flexibility to the final grid boxes to adapt the downscaling function. Overall, Figure \ref{fig:08} shows that the accuracy of Swin2SR outperforms DeepESD and UNet, with the exception of the mentioned area, and slightly over other high elevation regions.

In this following, we use the best-performing model resulting from the above intercomparison (Swin2SR) to compare the full-domain and the tiling approaches, the later providing an scalable and potentially cost-effective alternative, as described in Sec. \ref{subsec:tiling_approach}. The aim here is to evaluate the potential loss of performance of this method, caused by the division in tiles which pose the additional requirement of spatial generalisation and transferability. We compare both the tiling and patch variants, the former using fixed tiles covering the area of interest (see Figure \ref{fig:03}) and the latter using overlapping patches distributed randomly over the area. 

Table \ref{tab:performance_metrics_patches} shows the results for the different Swin2SR variants, the full-domain (Swin2SR-F), the tiling (Swin2SR-T), and the patch (Swin2SR-P) approaches. The later two exhibit similar errors with overall lower performance compared to its full-domain version, in particular RMSE and MAE values. For example, the RMSE for Swin2SR increases from 0.92 (full-domain) to 0.98 in the tile and patch configurations. The other error metrics, Bias, SSIM, and PSNR, exhibit similar results than the full-domain approach. Despite their overall lower performance of the Swin2SR tiling approach as compared to the full-domain results, this method still achieves similar results to other full-domain benchmarks, such as DeepESD. It is therefore, a cost-effective implementation when scalability is required.

\begin{figure}[H]
    \centering
    \includegraphics[width=0.82\linewidth]{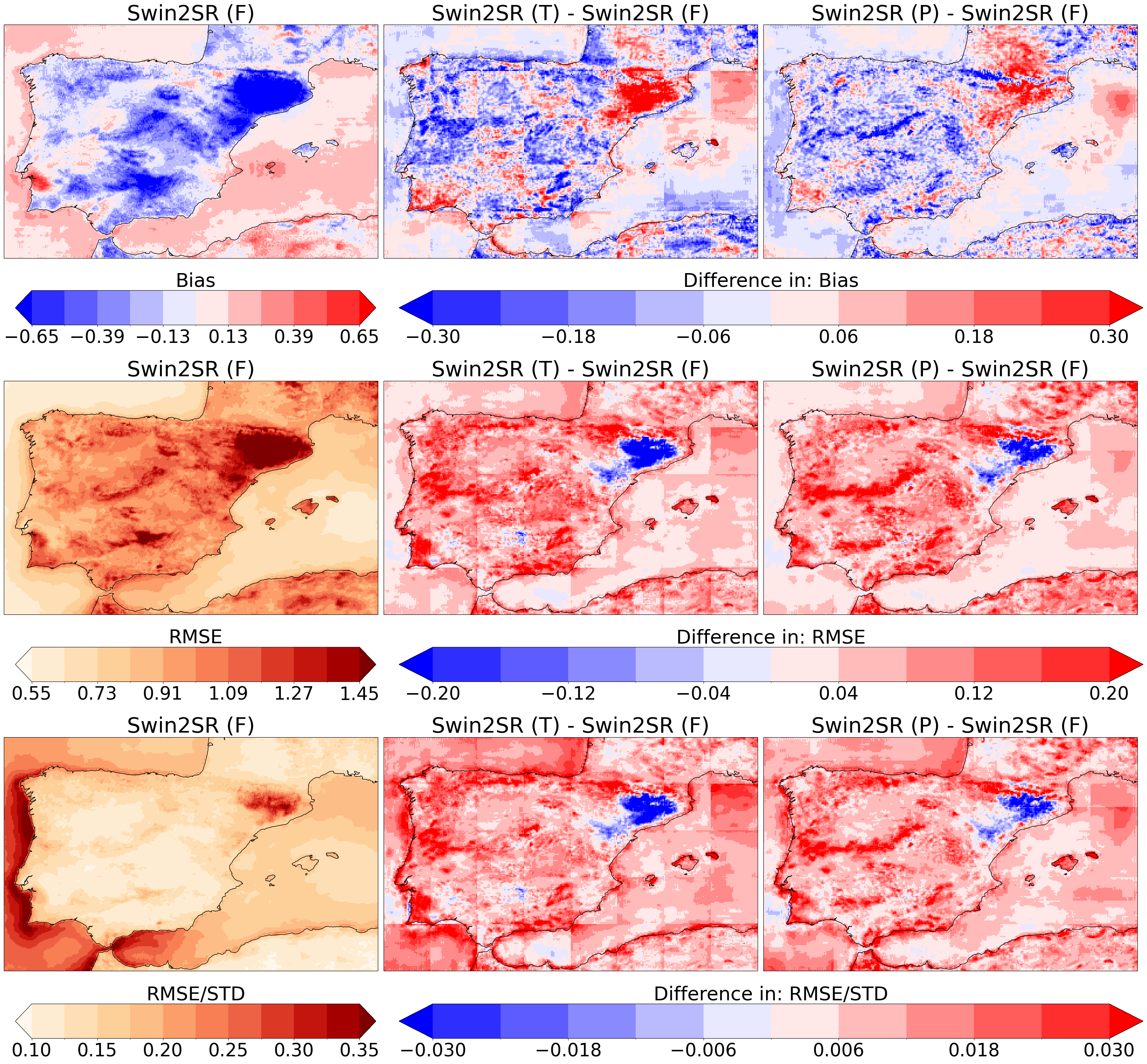}
    \caption{As Figure \ref{fig:08}, but comparing the Swin2SR in the full-domain approach (Swin2SR (F)) with the same model structure for the two patch-based implementations, using non-overlapping tiles (Swin2SR (T)), and using patches (Swin2SR (P)), in the second and third columns}
    \label{fig:09}
\end{figure}

Figure \ref{fig:09} shows the spatial errors for these methods. It can be shown that the tiling approach produces artefacts at the boundaries of the tiles, which are not exhibited by the patch version. Although the tiles include some spatial context (see the larger predictor regions in Figure \ref{fig:01}), this does not entirely eliminate the discontinuities. The patch approach accounts for that by increasing the variety of patches (selecting them randomly over the region) thus avoiding boundary artefacts. 

\begin{table}[h!]
\centering
\begin{tabular}{|c|r|ccccc|}
\hline
\rowcolor[HTML]{EFEFEF} 
\multicolumn{1}{|l|}{\cellcolor[HTML]{EFEFEF}}          & \multicolumn{1}{l|}{\cellcolor[HTML]{EFEFEF}} & \textbf{DJF}   & \textbf{JJA}   & \textbf{MAM}   & \textbf{SON}   & \textbf{Annual} \\ \hline
                                                        & \textbf{Swin2SR-F}                            & \textbf{0,91}  & \textbf{0,96}  & \textbf{0,89}  & \textbf{0,89}  & \textbf{0,92}   \\
                                                        & \textbf{Swin2SR-T}                            & 0,99           & 1,03           & 0,95           & 0,95           & 0,98            \\
\multirow{-3}{*}{\textbf{RMSE}}                         & \textbf{Swin2SR-P}                            & 0,99           & 1,03           & 0,95           & 0,95           & 0,98            \\ \hline
\rowcolor[HTML]{EFEFEF} 
\cellcolor[HTML]{EFEFEF}                                & \textbf{Swin2SR-F}                            & \textbf{0,68}  & \textbf{0,72}  & \textbf{0,66}  & \textbf{0,67}  & \textbf{0,68}   \\
\rowcolor[HTML]{EFEFEF} 
\cellcolor[HTML]{EFEFEF}                                & \textbf{Swin2SR-T}                            & 0,74           & 0,78           & 0,71           & 0,71           & 0,73            \\
\rowcolor[HTML]{EFEFEF} 
\multirow{-3}{*}{\cellcolor[HTML]{EFEFEF}\textbf{MAE}}  & \textbf{Swin2SR-P}                            & 0,73           & 0,77           & 0,71           & 0,72           & 0,73            \\ \hline
                                                        & \textbf{Swin2SR-F}                            & \textbf{-0,02} & \textbf{-0,10} & \textbf{-0,02} & \textbf{-0,01} & \textbf{-0,04}  \\
                                                        & \textbf{Swin2SR-T}                            & -0,05          & -0,09          & -0,04          & -0,04          & -0,06           \\
\multirow{-3}{*}{\textbf{Bias}}                         & \textbf{Swin2SR-P}                            & -0,06          & -0,09          & -0,04          & -0,04          & -0,06           \\ \hline
\rowcolor[HTML]{EFEFEF} 
\cellcolor[HTML]{EFEFEF}                                & \textbf{Swin2SR-F}                            & \textbf{0,87}  & \textbf{0,90}  & \textbf{0,91}  & \textbf{0,90}  & \textbf{0,89}   \\
\rowcolor[HTML]{EFEFEF} 
\cellcolor[HTML]{EFEFEF}                                & \textbf{Swin2SR-T}                            & 0,84           & 0,89           & 0,89           & 0,88           & 0,88            \\
\rowcolor[HTML]{EFEFEF} 
\multirow{-3}{*}{\cellcolor[HTML]{EFEFEF}\textbf{SSIM}} & \textbf{Swin2SR-P}                            & 0,84           & 0,89           & 0,89           & 0,89           & 0,88            \\ \hline
                                                        & \textbf{Swin2SR-F}                            & \textbf{28,93} & \textbf{28,97} & \textbf{29,38} & \textbf{29,03} & \textbf{29,08}  \\
                                                        & \textbf{Swin2SR-T}                            & 28,24          & 28,42          & 28,81          & 28,71          & 28,55           \\
\multirow{-3}{*}{\textbf{PSNR}}                         & \textbf{Swin2SR-P}                            & 28,15          & 28,39          & 28,73          & 28,69          & 28,49           \\ \hline
\end{tabular}
\label{tab:performance_metrics_patches}

\caption{Performance comparison of Swin2SR for the patch-based approach across seasonal and annual metrics. This table presents the RMSE, MAE, Bias, SSIM, and PSNR values for the Swin2SR in the full-domain approach (Swin2SR-F) and the same model structure in the patches approach both with (Swin2SR-P) and without overlap (Swin2SR-T), evaluated across different seasons (DJF, JJA, MAM, SON) and annually.}
\end{table}

\begin{figure}[H]
    \centering
    \includegraphics[width=0.77\linewidth]{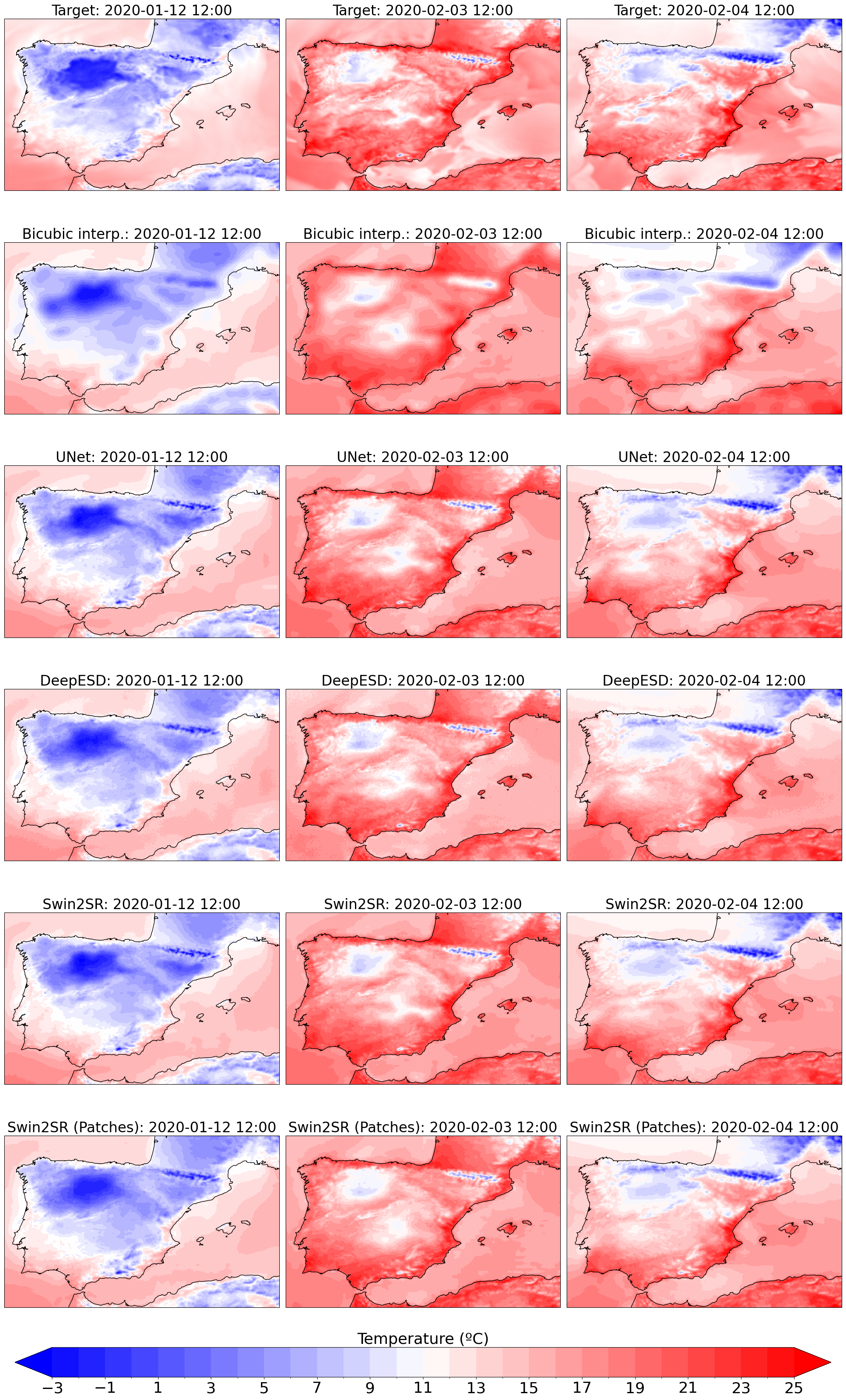}
    \caption{Spatial maps of the target (CERRA) data and predictions from different super-resolution methods for specific weather events in the Iberian Peninsula. Columns correspond to the dates of January 12th, February 3rd, and 4th, 2020, capturing periods of cold and warm anomalies. Rows represent results from bicubic interpolation, UNet, DeepESD, SwinSR, and SwinSR (Patches).}
    \label{fig:10}
\end{figure}

Finally, we conclude displaying the results of the different methods when applied to specific cases of winter cold  (January 13\textsuperscript{th}, 2020) and warm  (February 3\textsuperscript{rd} and 4\textsuperscript{th}, 2020) events across the Iberian Peninsula. High-resolution climate data from CERRA and the ERA5 bicubic interpolated counterparts, along with various super-resolution methods, are used to assess how each approach captures the spatial and temporal variability of these events, as shown in Figure \ref{fig:10}.

\section{Conclusions}
\label{sec:conclusions}

This study has conducted a comprehensive intercomparison of super-resolution methods for climate downscaling, demonstrating that Swin2SR is the most effective model when using the full-domain approach for both training and inference. Swin2SR consistently outperforms the second-best model, U-Net, across all metrics. For example, in terms of RMSE, Swin2SR achieves a relative improvement of around 8$\%$ annually (0.92 vs. 0.99), and the SSIM improvement is around 2$\%$ (0.89 vs. 0.87). These results highlight Swin2SR's ability to better capture fine-scale spatial patterns and minimise error, particularly in complex regions such as mountainous and coastal areas.

Furthermore, we explored the use of a tiling approach to improve model scalability, allowing for cost-effective continental-wide applications of these methods. This method conducts training over smaller tiles, which significantly reduces computational costs without greatly compromising accuracy. Although there is a slight increase in error (Swin2SR's RMSE rises from 0.92 to 0.99 in the tiling version) this degradation remains relatively small considering the substantial computational savings. The use of overlapping patches mitigates some of the boundary discontinuity issues and further improves performance. The patch-based approach presents a viable solution for large-scale applications, providing a trade-off between model performance and computational efficiency.

An important aspect of this study is the emulation of CERRA data using ERA5 as input. This emulation is especially valuable given that CERRA is not updated in real-time, with the most recent data available only up to 2021. In contrast, ERA5 is updated with a delay of approximately five days. Therefore, using our super-resolution methods, it is possible to emulate CERRA data with the same five-day delay as ERA5. This capability is crucial for providing higher-resolution climate data in near real-time, bridging the temporal gap in CERRA’s updates.

The SR approaches described in this paper were trained over the area of study on a V100 GPU with 16GB of memory. Note that for a continental-wide application the output would scale by a factor of 40 making it prohibitive for standard infrastructures. Therefore, the tiling approach provides a cost-effective alternative for scaling applications at the cost of an modest reduction of performance.

\section*{Acknowledgements}

This work extends the project initiated during the Code 4 Earth 2023 program, with support from experts at the European Centre for Medium-Range Weather Forecasts (ECMWF): Matthew Chantry, Andras Horányi, Mariana Clare, and Cornel Soci. Their guidance was essential for developing key aspects of this research.

We would like to thank our partners for their support throughout this work. This research was supported by the Horizon Europe project AI4EOSC ("Artificial Intelligence for the European Open Science Cloud"). We also acknowledge the European Weather Cloud for providing computing and storage resources, as well as expert support.

\bibliographystyle{apalike}  
\bibliography{references}

\end{document}